\documentclass[preprint,12pt,authoryear]{elsarticle}

\usepackage{amssymb}
\usepackage{makecell}

\usepackage{xcolor} 
\definecolor{myblue}{RGB}{0,153,255} 
\usepackage{booktabs}
\usepackage{url}
\journal{Image and Vision Computing}

\begin{document}

\begin{frontmatter}



\title{Occlusion-Aware Deep Convolutional Neural Network via Homogeneous Tanh-transforms for Face Parsing}


\author[inst1]{Jianhua Qiu}
\ead{qiujianhua@hnu.edu.cn}
\affiliation[inst1]{organization={Hunan University.},
            addressline={No.116 Lushan South Road}, 
            city={Changsha},
            postcode={410082}, 
            state={Hunan},
            country={China}}

\author[inst2]{Weihua Liu$^{\star}$}
\ead{liuweihua@a-eye.cn}

\affiliation[inst2]{organization={AthenaEyesCO., LTD.},
            addressline={Building 14, Zhongdian Software Park, No. 39 Jianshan Road}, 
            city={Changsha},
            postcode={410205}, 
            state={Hunan},
            country={China}}

\author[inst2]{Chaochao Lin}
\ead{linchaochao@a-eye.cn}
\author[inst2]{Jiaojiao Li}
\ead{lijiaojiao@a-eye.cn}
\author[inst3]{Haoping Yu}
\ead{hyu90@jh.edu}
\author[inst4]{Said Boumaraf}
\ead{said.boumaraf@ku.ac.ae}
\affiliation[inst3]{organization={Johns Hopkins University},
            addressline={3400 N Charles Street}, 
            city={Baltimore},
            postcode={21218}, 
            state={Maryland},
            country={United States}}

\affiliation[inst4]{organization={Khalifa University of Science and Technology},
            postcode={127788}, 
            state={Abu Dhabi},
            country={United Arab Emirates}}
 \fntext[]{\textsuperscript{*} Corresponding author: Weihua Liu}

\begin{abstract}
Face parsing infers a pixel-wise label map for each semantic facial component. Previous methods generally work well for uncovered faces, however, they overlook facial occlusion and ignore some contextual areas outside a single face, especially when facial occlusion has become a common situation during the COVID-19 epidemic. Inspired by the lighting phenomena in everyday life, where illumination from four distinct lamps provides a more uniform distribution than a single central light source, we propose a novel homogeneous tanh-transform for image preprocessing, which is made up of four tanh-transforms. These transforms fuse the central vision and the peripheral vision together. Our proposed method addresses the dilemma of face parsing under occlusion and compresses more information from the surrounding context. Based on homogeneous tanh-transforms, we propose an occlusion-aware convolutional neural network for occluded face parsing. It combines information in both Tanh-polar space and Tanh-Cartesian space, capable of enhancing receptive fields. Furthermore, we introduce an occlusion-aware loss to focus on the boundaries of occluded regions. The network is simple, flexible, and can be trained end-to-end. To facilitate future research of occluded face parsing, we also contribute a new cleaned face parsing dataset. This dataset is manually purified from several academic or industrial datasets, including CelebAMask-HQ, Short-video Face Parsing, and the Helen dataset, and will be made public. Experiments demonstrate that our method surpasses state-of-the-art methods in face parsing under occlusion.
\end{abstract}


\end{frontmatter}


\section{Introduction}
\label{sec:introduction}


Face parsing aims to predict labels for each pixel in input face images based on pixel semantics, and the parsed results are widely used for other applications, such as face analysis \citep{selfa2005place}, face editing \citep{de2015growth}, face swapping \citep{bitouk2008face} and face recognition \citep{zhao2003face}. 
While parsing a face is very similar to generic scene segmentation \citep{hariharan2014simultaneous, noh2015learning}, they are not the same because faces have similar contexts and translation invariant. With the rapid development of convolutional neural networks (CNN) \citep{lecun1989backpropagation}, a plethora of methods for basic face parsing tasks has been widely developed due to the position-independent characteristic. \citep{yin2021end}. 

Previous face parsing methods have encountered challenges in effectively addressing occlusions. Based on their modeling approaches, classical face parsing methods are typically categorized into two main types: methods that consider the entire face as a holistic entity, and methods that focus on modeling individual facial components separately.
The former examines an entire face \citep{liu2015multi}, whereas the latter models each facial component individually. \citep{liu2017face, luo2012hierarchical}. 
\citet{lin2019face} ’s experiment shows that the former methods can label the background and hair better, but their performance is worse than latter methods and they mostly process the face in Cartesian coordinates system.  The recent emergence of non-Cartesian coordinate systems provides an alternative to the constraints of classical methods.
\citet{liu2017face} and \citet{lin2019face} perform the face parsing tasks in a new coordinate system and achieve significant improvement.
However, when key face components, such as the nose and mouth, are occluded, the performance of these methods suffers.
Because of the COVID-19 pandemic, the number of people wearing masks to cover their faces has increased significantly \citep{wang2020association}, making occluded face parsing difficult.
Sunglasses, eye masks, and fake noses are also examples of facial occlusions.
In addition, sunglasses, eye masks, and fake noses are examples of facial occlusions.
As a result, when analyzing occluded faces, classical methods are prone to misidentify occlusions as face components with similar positions, resulting in unsatisfactory results.

The main cause of the aforementioned issues is summarized, which is that previous methods frequently failed to pay attention to the edges of the face.
As a result, our method attempts to focus on the four corners of a face, drawing inspiration from image illumination theory.
More facial details can be captured with the illumination theory when divide a face image into four parts and perform a kind of transform into a non-Cartesian coordinate system, because this is equivalent to dispersing a center point light source into an area light source. 
The four slices of face details can be enhanced separately by this process, and occlusions have a limited effect on all slices.
This process can improve the four slices of face details individually, and occlusions have a limited effect on all slices. 
Based on this characteristic, we propose a novel homogeneous Tanh-transforms, called four-point transform, which transforms the image to the four Tanh-polar coordinate systems using the four corner points of the face as the origin respectively. 
Tanh-polar representation differs from the original Cartesian representation and it contains the contextual information of the face image. 
Then, the transformed face is fed into an occlusion-aware convolutional neural network. 
A new building block, called Four-point block (FPB) and a special occlusion-aware loss are designed to better extract occluded features.
When compared to existing face parsing method, the proposed architecture improves the robustness of face parsing under occlusion and can be trained in an end-to-end fashion. 
In order to improve the anti-occlusion capability of our model and to validate the superiority of the method in facial occlusion, we present Sheltered Face Parsing Dataset which contains 52,097 in-the-wild images. 
The annotation categories are reorgnized for these images to include 11 face regions and 3 occlusions. Extensive experiments show that these approaches have strong anti-occlusion capabilities, as well as excellent performance for both intra-dataset and cross-dataset evaluation.

\begin{figure}[h]
	\centering
	\includegraphics[width=5in]{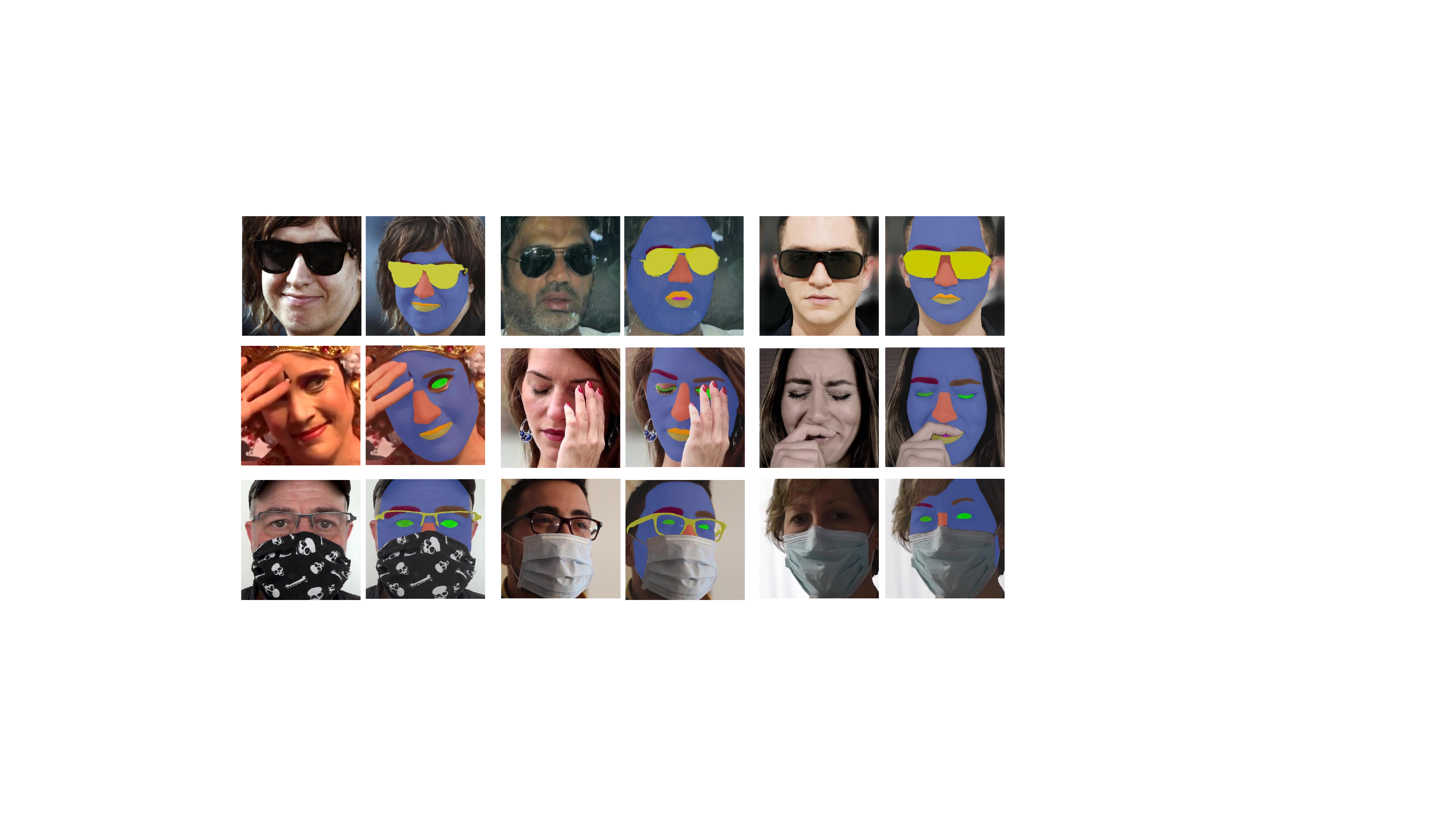}
	\caption{Examples from our Sheltered Face Parsing Dataset and the color-coded labels predicted by our proposed method, which performs precise parsing of each facial component when some objects cover the face.}
	\label{fig_1}
\end{figure}

Fig. \ref{fig_1} shows the benefit of the proposed method on a few samples from the test set. Our method is insensitive to various forms of occlusions and more importantly, arrive at much stable predictions that hard to attain with ordinary coordinate.

Overall, our contributions are summarized as follows:

\begin{itemize}
    \item A novel four-point transform neural network (FTNet) is proposed to address the problem of face parsing under occlusion which takes full advantage of the contextual information of the face image and report experimental results in face parsing dataset. To better elucidate the motivation and principles of our method, we used lighting scenarios to illustrate, akin to illumination theory. As far as we know, this serves as an intriguing analogy.
    \item The homogeneous Tanh-transforms, called four-point transform, is introduced, which is used to warp faces into the Tanh-polar coordinate system by four corner points of expanded bounding box. In Tanh-polar coordinate systems with four points as the origin respectively, the key information of the facial components is enhanced which can improve the generalization capability of occluded face parsing.
    \item Leveraging the proposed homogeneous Tanh-transforms, the structure of basic convolutional neural network with Four-point Block (FPB) is designed for occluded faces. Using our coordinate transform as constraints, a new loss function named occlusion-aware loss is created, which enhances occlusion segmentation. Our overall approach is simple and can be trained end-to-end.
    \item As an additional contribution, the Sheltered Face Parsing Dataset is presented, which is a novel in-the-wild sheltered face parsing dataset with over 54 thousand images and we intend to make public to aid future research.
\end{itemize}

The remainder of the paper is organized as follows. Section \ref{sec:related} discusses related work, Section \ref{sec:methodology} describes our proposed method, Section \ref{sec:experimental} reports the experimental result and Section \ref{sec:conclusion} concludes our contribution along with future work.

\section{Related Work}
\label{sec:related}
Face parsing research has yielded impressive results, thanks in large part to advances in deep learning. This section provides a brief overview of related work on face parsing, image polar transform, and occlusion handling.
\subsection{Face Parsing}
Most existing approaches for face parsing can be categorized into two groups: global-based and local-based. 

Global-based methods analyze the entire image and directly predict per-pixel semantic labels. Earlier works concentrated on developing a model to represent the spatial relationships of entire images \citep{warrell2009labelfaces, smith2013exemplar}. With the rapid development of CNN, face parsing models tend to use network designs with multiple loss functions \citep{liu2015multi}. \citet{zhou2017face} created a unified framework by combining FCN, super-pixel information, and the CRF model. Nonetheless, \citet{lin2019face} ’s experiment showed that these global-based approaches can only provide a limited accuracy due to the lack of focus on facial components. Local-based methods train the model for each facial component separately in order to predict the mask for each part individually. Early works on local-based methods predicted all facial components separately, then combined them based on spatial correlations \citep{luo2012hierarchical}. \citet{liu2017face} improved accuracy using a two-stage approach that included a shallow CNN and a spatially variant recurrent neural network. ICNN \citep{zhou2015interlinked} incorporates two stages of face parsing, first locating the facial component and then labelling the pixels of the identified facial parts. Local-based methods can improve facial component accuracy, but their accuracy in predicting hair and background is limited. 

\subsection{Image Polar Transform}
Most existing computer vision models are built on images in a Cartesian coordinate system. These models are very sensitive to many complex affine transformations such as rotation and scaling due to the large number of irreversible interpolations required. Therefore, many image processing algorithms that convert the original image to a representation in another coordinate system have been investigated. Among them, polar coordinate is a well-known sampling method in the field of image processing. 

\citet{salehinejad2018image} used a sampling method based on radial transformation to augment training data, which transform each input image into multiple images in polar coordinate systems, thus facilitate the training of models from limited source data.  \citet{kim2018cnn} proposed a user-guided point as the origin of the polar coordinate system. The transformed image is then segmented using a CNN. \citet{jiang2019polar} created a novel polar coordinate CNN (PC-CNN) for rotation invariant feature learning, which transformed the rotation invariance of training samples into translation invariance through polar coordinate transform. \citet{fu2018joint} introduced a polar transform to represent the original image in a polar coordinate system, which further improved the segmentation performance of the proposed architecture.
Furthermore, \citet{lin2019face} used the RoI Tanh-Warping to alter the image's coordinate system, which transformed the face in the image's center. The performance was then improved by \citet{lin2021roi} by proposing a Tanh-polar representation and introducing a well-designed CNN backbone. However, unlike our proposed method, these methods ignore the impact of occluded scenes on the face parsing task, rendering them ineffective in the presence of occlusion.

\subsection{Occlusion Handling}
Occlusion handling strategies have been extensively studied. 
\citet{gao2011segmentation} proposed a segmentation-aware model with binary variables to handle occlusions, which denote the visibility of objects in each bounding box cell. 
\citet{ghiasi2014parsing} used deformable models with local templates to handle occlusion in human pose estimation. 
\citet{chen2015multi} used an energy minimization framework to address the occlusion problem by incorporating top-down class-specific inference and shape prediction as well as bottom-up segments. 
\citet{zheng2020novel} proposed an effective feature extraction approach to capture the local and contextual information for occluded face recognition. 
\citet{ke2021deep} addressed occlusion through explicit modelling of occlusion patterns in shape and appearance. 
\citet{albalas2022learning} developed a model to detect occluded or masked faces based on fused spatial-based convolutional graphs of key facial features.
However, they only focus on the occlusions in other natural scenes, rather than the more difficult occlusions on the face.

\section{Methodology}
\label{sec:methodology}
This paper introduces a four-point transform network (FTNet) with homogeneous Tanh-transforms for occluded face parsing. The homogeneous Tanh-transforms perform four Tanh-polar transforms based on four points of an image, so we also call it four-point transform, which will be explained in section \ref{sec:3.1}. Given a face image of Cartesian coordinate and a bounding box of the target face, four-point transform warps the entire image into four Tanh-polar coordinate systems and stitches them together. The proposed four-point transform focuses on the non-occluded area of the target face to extract more robust spatial features under occlusion.

\begin{figure}[h]
	\centering
	\includegraphics[width=5in]{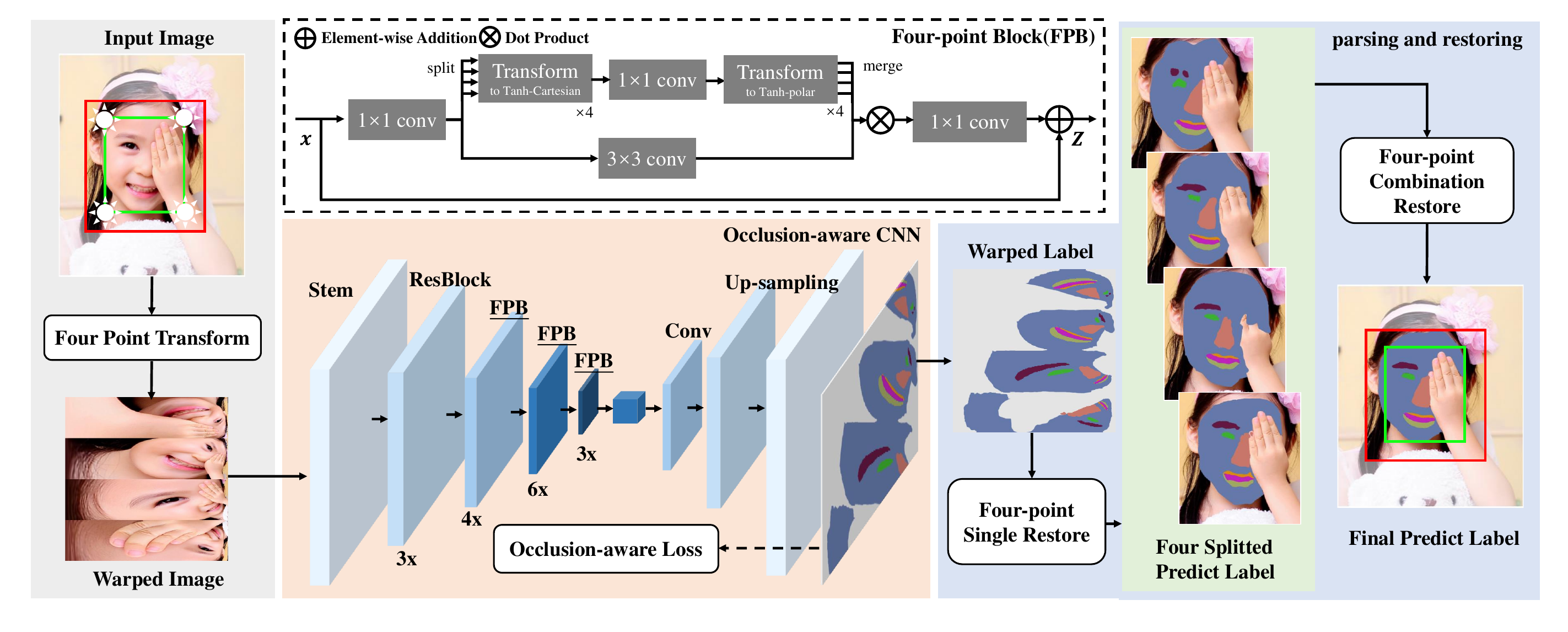}
	\caption{The overview of proposed FTNet structure. Given an input image and bounding box (in green), the four-point transform is applied to convert the RoI (in red) of input image to a new representation. The warped origin of input image (in white light) is defined as the corner of bounding box. The proposed occlusion-aware CNN is mainly composed of multiple Four-point Block (FPB) detailed in \ref{subsec:feature} to extract feature from the warped image. An occlusion-aware loss is designed to guide the warped parsing label prediction. Finally, the warped label is restored to the regular coordinate system. }
	\label{fig_2}
\end{figure}

The whole pipeline of our method is shown in Fig. \ref{fig_2}. Our framework consists of four major components: 
1) the four-point transform, which warps input face image of Cartesian coordinate into occlusion-aware Tanh-polar coordinate; 
2) the backbone feature extraction module, which extracts informative features from the warped face for subsequent operations; 
3) the parsing and restoring modules, which predict the segmentation scores through up-samplings on the feature map and restore the result into Cartesian coordinate; 
4) the occlusion-aware loss, which introduces a penalty coefficient for the false prediction of occlusions. 
Next, each part will be introduced in detail.

\subsection{Homogeneous Tanh-transforms and Restore}
\label{sec:3.1}
Previous face parsing works typically used face alignment techniques, which resulted in peripheral information loss \citep{liu2015deep}. 
This is due to large variations in the target face's position, rotation, and size, and the network's input typically ignores the importance of the uncovered facial components. Traditional FCN can handle a large portion of face parsing, but its accuracy is limited, particularly when occlusion is present, due to a lack of focusing on each non-occluded individual part. To solve this problem, we propose homogeneous Tanh-transforms, called four-point transform, which warps the target face into Tanh-polar coordinate through four focusing points and makes precise parsing. Our warping approach transforms the entire image to the multiple homogeneous Tanh-polar representation, which expands the ratio of the region of interest (RoI) and compresses the background information, which lead to less data loss \citep{lin2019face, lin2021roi}. 

\begin{figure}[h]
	\centering
	\includegraphics[width=5in]{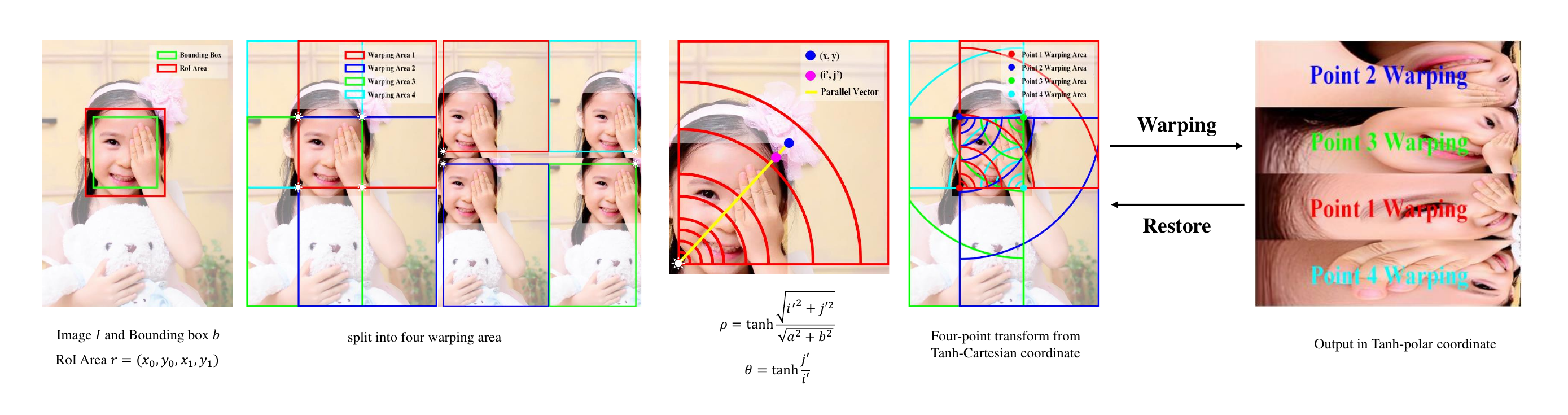}
	\caption{The homogeneous Tanh-transforms pipeline, namely, four-point transform. (1) Input image, bounding box and the RoI area (2) Four warping regions according to four corners of bounding box (3) An example of the homogeneous Tanh-transforms in one warping area (4) The image in Tanh-Cartesian coordinate system and warping output in Tanh-polar coordinate system. Note that semi-transparent area which is out of RoI area do not participate in the calculation.}
	\label{fig_3}
\end{figure}

Fig. \ref{fig_3} illustrates the process of the homogeneous Tanh-transforms. Given an image $I$, as shown in Fig. \ref{fig_3}, the RoI rectangle $r=(x_1',x_2',x_3', x_4')$ is presented by multiplying the size of the bounding box $b=(x_1,x_2,x_3,x_4)$ by a variable zoom factor $l$ and all subsequent calculations are performed in the RoI region. Following the concept of warping, our method chooses four points as the original warping points, which chosen from the corner of bounding box $b$. These points divide the entire image into four warping regions for four-point transform. According to the bounding box $b=(x_1,x_2,x_3,x_4)$ , calculate the width and height of $r$: $w=x_2-x_1,h=y_2-y_1$. Then, the image that warped by four-point transform can be defined as:
$$I^{'}=FT(I,r)=P(W(I_i,I_j,w,h)),\forall I_i\in I_x, \forall I_j\in I_y$$
where $I'$ is the warped image, $FT$ is the function of four-point Transform, $P$ is a bilinear interpolation function which can resample discrete coordinates of an image and project them onto determined coordinates system linearly by using a typical interpolation method \citep{jaderberg2015spatial, tai2019equivariant}, and $W$ is the mapping function that projects every pixel to the Tanh-polar coordinate.

In general, the center of bounding box is considered as the original point. However, to determine the mapping function of four-point transform $W(i,j,w,h)$ for each pixel $(i,j)$, it first need to transform the pixel position to make sure the original points are separately located at the four corners of bounding box. For an example, in the warping area 1 which highlighted in red in Fig. \ref{fig_3}, the relative position  $(i',j')$ of (i,j) is defined as:
$$i^{'}=i-\frac{i}{\left|i\right|}\frac{w}{2}, j^{'}=j-\frac{j}{\left|j\right|}\frac{h}{2}$$  

As Fig. \ref{fig_3} shows, the vector $(x,y)$ is parallel to the vector $(i',j')$, and $(x,y)$ is described by the canonical equation of the ellipse $\frac{x^2}{a^2}+\frac{y^2}{b^2}=1$, whose major axes $a=\frac{w}{2\sqrt{\pi}}$ and minor axes $b=\frac{w}{2\sqrt{\pi}}$ are related to the height/width of the bounding box. Therefore, $x$ and $y$ satisfy the following equation: 
$$x^2=\frac{(\frac{w}{2\sqrt{\pi}})^2\cdot((\frac{h}{2\sqrt{\pi}})^2-y^2)}{(\frac{h}{2\sqrt{\pi}})^2}=\frac{w^2}{h^2}(\frac{h^2}{4\pi}-y^2)$$

Then, the Tanh-polar coordinate system by mapping function is defined as:
$$W(i,j,w,h)=(\theta, \rho)=(\rm{tanh}(\frac{j'}{i'}),\rm{tanh}(\frac{\sqrt{i'^2+j'^2}}{\sqrt{a^2+b^2})})$$
where $\theta$, $\rho$ are the polar angle and diameter, respectively.

For the other three warping regions, the homogeneous Tanh-transforms are the same as described above. The only difference is the relative position $(i',j')$ when different corners are used as the origin of the coordinate system. Four transforms on the four warping regions constitute our four-point transform method. As shown in Fig. \ref{fig_3}, the image $I$ is warped into four parts after this mapping algorithm and then merged together. It maps each original Tanh-Cartesian coordinate system pixel to four Tanh-polar coordinate systems and then interpolates unmapped pixels. Each part focuses on one corner of the bounding box, and its area takes up the majority of the image. Though our homogeneous Tanh-transforms, fixed-size images with higher proportions of RoI regions can be produced. The transformed image contains all information inside the bounding box, and some losses occur outside of the bounding box, while enhancing the compressed information of non-RoI regions for use in feature extraction modules.

For restore, it first recovers four split labels from four parts, then calculates and combines them based on a selected basis. However, because some duplicated warping areas exist in more than one part, as shown in Fig. \ref{fig_3}, it is necessary to perform the following step to restore the transformation, which restore four individual sections that correspond to four parts in
$$I=FT^{-1}(I^{'},r)=P(W_{combine}^{-1}(W^{-1}(I_i^{'},I_j^{'},w,h))),\forall I_i^{'}\in I_x^{'},\forall I_j^{'}\in I_y^{'}$$
where $W_{combine}^{-1}$ combines these four sections, then the selected basis calculates and combines overlapped areas. In our system, the mean function is chosen as our base calculation.

\begin{figure}[h]
	\centering
	\includegraphics[width=5in]{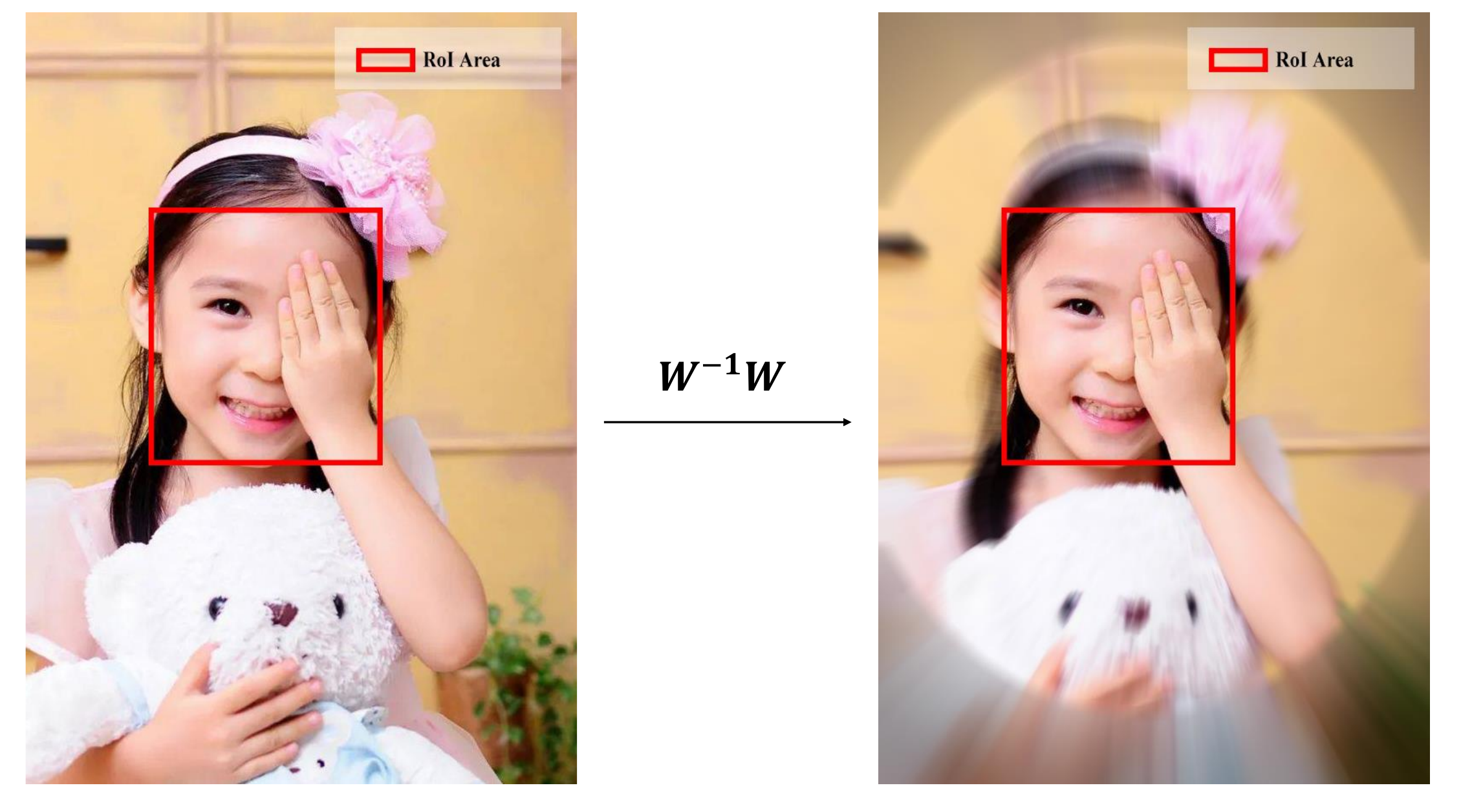}
	\caption{Comparison of original image and restored image. Observing that outside the RoI area, some information losses.}
	\label{fig_4}
\end{figure}
 
Fig. \ref{fig_4} provides the comparison between the original image $I$ and the restored image $W^(-1) W(I)$ . As the figure illustrates, we can observe that $W^(-1) W(I)$  restores most of the information in the RoI area, and other areas are basically recovered. Although some losses happen in the areas outside of RoI, the losses are limited since most important labels are inside the RoI area for face parsing.

\subsection{Four-point Block}
\label{subsec:feature}
Given the warped face image $W(I,r)$, which contains the compressed information of whole image through the four-point transform, the backbone feature extraction modules are deployed to capture implicit features under Tanh-polar coordinate. 
However, although the information of the rotation equivariance can be preserved, four-point transform may face a risk of losing the translation equivariance.
Therefore, we upgrade the original residual block of feature extraction to a new building block, called Four-point Block. 
It is able to extract features more efficiently in a Tanh-polar coordinate.

Fig. \ref{fig_2} illustrates the branch structure of the FPB in detail. 
The four-point branch in Tanh-polar coordinate system learns the rotation equivariant representation. The Tanh-Cartesian branch, which transforms feature maps into Tanh-Cartesian coordinate systems, learns the translation equivariant representation, so as to achieve the fundamental purpose of complementary learning. The Tanh-Cartesian coordinate system can be defined by
$$TC(\vec{v}):=\rm{tanh}(\frac{i'}{\sqrt{x^2+y^2}})\rm{tanh}(\frac{j'}{\sqrt{x^2+y^2}})$$
where $TC(i,j)$ is the function of Tanh-Cartesian transform.

The input to FPB is a four-point representation $X_{fp}$ of shape $(h,w,c)$. 
The residual path uses a stack of $1\times 1$, $3\times 3$ and $1\times 1$ convolutions following the bottleneck design \citep{he2016deep}. 
The first $1\times 1$ conv layer is used to reduce the channel dimension. Its output feature maps are divided into two branches as described above. In Tanh-Cartesian branch, the feature maps are split back to four four-point warping regions and transformed to the Tanh-Cartesian space, respectively. In Tanh-cartesian coordinate system, four $1\times 1$ conv layers are used to compute these four-point feature maps, which then are transformed to the Tanh-polar space through four-point transform and merged together. In four-point branch, a $3\times 3$ conv layer is used to compute feature maps in Tanh-polar space. Then, the feature maps from these two branches are multiplied in Tanh-polar coordinate system. The last $1\times 1$ conv layer restores the channel dimension so the residual representation can be added to the input $X_{fp}$. 

\subsection{FTNet Architecture}
Through homogeneous Tanh-transforms, namely, four-point transform, the input image of the Tanh-polar coordinate system is obtained which compresses more RoI key information and surrounding semantics. For a higher quality feature extraction module fusing the advantage of Tanh-Cartesian and Tanh-polar representation, we propose FPB for composing our new backbone network, four-point transform network (FTNet). FPBs are stacked according to the structure of ResNet to build FTNet for feature extraction. Thanks to the grouped conv $1\times 1$ and the conv $3\times 3$ with halved channels, the overall number of parameters are less than the ResNet50 backbone.
To predict the masks, we use the FCN-8's \citep{long2015fully} decoder. More advanced decoders necessitate dedicated hyperparameter tuning, which can significantly affect performance. To map the feature maps to pixel-wise prediction logits, the decoder employs two conv $3\times 3$ layers and a bilinear up-sampling layer. Each component of the transformed image is invertible in theory.

\subsection{Occlusion-aware Loss}
Based on homogeneous tanh-transforms, we design an occlusion robust network as described above, namely FTNet. The focus of the loss function used in the current face parsing field is mainly on the classification of pixels, while ignoring the adverse effects of masks, glasses and other occlusions. Therefore, to train our FTNet, an especial occlusion-aware loss is introduced to further address the face occlusion problem in face parsing. Furthermore, in order to better focus on occlusion semantic information, different weights are assigned to occluded pixels and non-occluded pixels in contrast to treating all pixel equally as the shape of the parsed face will be destroyed by occlusion. So, the loss weight of the pixels in the occluded face region are increased which added as the constraints of the loss function as follows,

$$L_{FTNet}=l\cdot L_{ce}+(1-l)L_{dice}+\alpha \frac{1}{N}\sum_i P(g_i^f)||g_i^f-s_i^f||_2^2$$
where
$$L_{ce}=\frac{1}{N}\sum_i\sum_c g_i^c \log{s_i^c}$$
$$L_{dice}=1-\frac{2\sum_i\sum_c g_i^cs_i^c}{\sum_i\sum_c {g_i^c}^2+\sum_i\sum_c {s_i^c}^2}$$

Here, $N$ is the total training samples of a batch, $\alpha$ is the hyperparameters of constraints, $g_i^c$   denotes the ground truth of class $c$ in $i$-th sample, $s_i^c$   denotes the predicted probability of class $c$ in $i$-th sample, $f$ denotes the face skin class, $P (g_i^f)$ is the occlusion value of $g_i^f$ calculated from the mean of itself and the surrounding 8 pixels ($P (g_i^f)$ is high when $g_i^f$ is occluded) and the weight $l$  is determined by the average point to curve Euclidean distance among points around curve of predicted segmentation to the ground truth.

\section{Experimental Evaluation}
\label{sec:experimental}
\subsection{Implementation Details}
\label{subsec:4.1}
\textbf{Dataset}. 
We perform experiments on a new dataset called Sheltered Face Parsing Dataset (SFPD). SFPD has 52097 images, collected from CelebAMask-HQ \citep{lee2020maskgan}, the Short-video Face Parsing (denoted as SVFP) \citep{cvpr2021}, and the HELEN dataset \citep{le2012interactive} which contains facial occlusion challenges. Table \ref{tab_1} compares the number of different datasets and illustrates that SFPD contains in-the-wild images and more occluded images which is of higher difficulty to face parsing. The collected images are carefully screened for no watermarks or blurring, and manually reannotated to eliminate label discrepancies. Furthermore, SFPD is more concerned with facial occlusions which presented by elaborate pixel-wise annotation, including 9 general face part categories and 4 additional face occlusion categories (i.e., hand occlusion(hand\_occ), eyeglasses(e\_glasses), sunglasses and mouth mask(m\_mask)).The images are divided into 42097 samples for training, 5000 for validation and 5000 for testing, respectively. 

\begin{table}[!htbp] 
\centering
\begin{tabular}{cccccc} 
\toprule 
\multicolumn{3}{c}{Benchmark}& images& In-the-wild & occluded images\\  
\hline 
\multicolumn{3}{c}{CelebAMask-HQ}& 27176& $\times$ & 828\\   
\multicolumn{3}{c}{SVFP}&19445&\checkmark& 974\\
\multicolumn{3}{c}{Helen}&2000&\checkmark& 60\\
\multicolumn{3}{c}{Ours}&42097&\checkmark& 1381\\
\bottomrule 
\end{tabular}
\caption{The comparison of training set of face parsing datasets.}
\label{tab_1}
\end{table}

\textbf{Training}. 
Our method is implemented on PyTorch and trained on two Tesla V100S GPUs. For data preparation, we apply data augmentation techniques for all the training samples to avoid over-fitting, e.g., random rotation and scale augmentation. The rotation angle is randomly selected from $(-30^{\circ}, 30^{\circ})$, while the scale factor is randomly selected from $(0.75, 1.25)$. During training, our four-point transformed images are warped from original images and resized to $512\times 512$. For our proposed occlusion-aware loss, the hyperparameters $\alpha$ is set to 2. For optimization, we adopt the Stochastic Gradient Descent (SGD) and we use the poly learning rate schedule \citep{chen2017deeplab}, $lr^*=(1-\frac{iter}{max\_ iter})^p$, $p=0.9$, in which the initial $lr$ is $0.01$. The $max\_ iter$ is  $epochs\times batch size$, where $batch size=8$ and $epochs=50$.

\textbf{Evaluation}. 
All the testing procedures are carried out on a single Tesla V100S GPU. For each test sample, the size of the predicted labels is set to $512\times512$ and scaled to original size for visualization. For evaluation, we report F1 score (F1) for all categories and their mean. The predicted masks are evaluated on the original image scale. The F1 can be represented by the formula as follow:
$$F=\frac{2\rm{Precision}\cdot\rm{Recall}}{\rm{Precision}+\rm{Recall}}$$

\subsection{Comparisons}
\textcolor{black}{The thorough comparisons are performed between our model and 6 state-of-the-arts (i.e., \citet{masi2020towards}, \citet{lin2021roi}, \citet{huang2021clrnet}, \citet{kim2022end}, \citet{li2023mask}, \citet{han2023scanet}) on our dataset in Tab.\ref{tab_2}. \citet{lin2021roi} is related to our work which also adopt image transform and \citet{li2023mask} propose a de-occlusion module for parsing occluded faces in a semi-supervised fashion, while other methods \citep{masi2020towards, huang2021clrnet, kim2022end,han2023scanet} are currently performing well in face parsing.The majority of the methods referenced are open-source. If not available as open-source, we implemented the models ourselves, with appropriate adjustments based on the literature's hyperparameters.These methods focus on the related to the shape of facial components to improve the face parsing performance, in detail, including consistent based learning mechanism \citep{ masi2020towards}, and attention based feature extraction \citep{huang2021clrnet, kim2022end,han2023scanet}. Results are measured by F1 which is commonly used by existing face parsing literature. We present the comparison results for both non-occluded and occluded images from the SFPD dataset in Tab.\ref{tab_2} and Tab.\ref{tab_2.2} separately. Each column shows the F1 percentage corresponding to a specific face label, including our reannotation and the overall F1-score as well. }

\begin{table}[!t] 
\centering
\begin{tabular}{ccccccccccc} 
\toprule 
\multicolumn{3}{c}{Methods}& skin& l\_brow & r\_brow &l\_eye &r\_eye  \\  
\hline 
\multicolumn{3}{c}{\citet{masi2020towards}}&91.4 &89.16 &85.21 &92.07 &91.39 \\
\multicolumn{3}{c}{\citet{lin2021roi}}& \textbf{94.22} & 90.70  & 90.10 &\textbf{95.59} &94.84\\   
\multicolumn{3}{c}{\citet{huang2021clrnet}}&90.87  &89.23  &85.13 &90.14  &90.72 \\
\multicolumn{3}{c}{\citet{kim2022end}}&92.89 &91.20 &91.05 &91.18 &\textbf{95.77} \\
\multicolumn{3}{c}{\citet{li2023mask}}&91.73 &90.24 &90.32 &89.78 &89.94 \\
\multicolumn{3}{c}{\citet{han2023scanet}}&91.25 &92.82 &92.46 &88.79 &89.17 \\
\hline 
\multicolumn{3}{c}{FTNet(Ours)}&92.04 &\textbf{94.31} &\textbf{93.26} &92.71 &91.50 \\
\bottomrule 
\toprule 
\multicolumn{3}{c}{Methods} &nose&u\_lip &l\_lip&inner\_mouth &\textbf{overall} \\  
\hline 
\multicolumn{3}{c}{\citet{masi2020towards}}&86.97 &83.98 &89.81 &91.63 &89.07 \\
\multicolumn{3}{c}{\citet{lin2021roi}}& 90.25 & 90.96 & 84.28 &85.12 &90.67 \\   
\multicolumn{3}{c}{\citet{huang2021clrnet}}&88.48   &83.39  &89.47 &91.07 &88.72 \\
\multicolumn{3}{c}{\citet{kim2022end}}&86.62 &90.06 &83.74 &\textbf{96.14} &90.96 \\
\multicolumn{3}{c}{\citet{li2023mask}}&90.64 &89.37 &89.52 &92.17 &90.41 \\
\multicolumn{3}{c}{\citet{han2023scanet}}&89.82 &90.23 &90.11 &89.87 &90.50 \\
\hline 
\multicolumn{3}{c}{FTNet(Ours)}&\textbf{93.97} &\textbf{92.56}&\textbf{92.60} &92.22 &\textbf{92.80} \\
\bottomrule 
\end{tabular}
\caption{Comparison with state-of-the-art methods on non-occluded images from the SFPD dataset. F1 scores are represented in percentage.}
\label{tab_2}
\end{table}
\vspace{8mm}

\begin{table}[!t] 
\centering
\begin{tabular}{cccccc} 
\toprule 
\multicolumn{1}{c}{Methods} &h\_occ &e\_glasses &sunglasses &m\_mask &\textbf{overall} \\  
\hline 
\multicolumn{1}{c}{\citet{masi2020towards}}& 89.18 & 88.09 &84.37 &90.15 &87.95 \\
\multicolumn{1}{c}{\citet{lin2021roi}}&\textbf{91.14} & 89.75 & 91.16 & 84.25 & 89.08 \\   
\multicolumn{1}{c}{\citet{huang2021clrnet}}& 89.46 & 88.46 &84.96 &90.22 &88.28  \\
\multicolumn{1}{c}{\citet{kim2022end}}& 90.88 & 90.59 &91.68 &85.45 &89.65 \\
\multicolumn{1}{c}{\citet{li2023mask}}&90.87 &89.46 &91.29 &90.81 &90.61 \\
\multicolumn{1}{c}{\citet{han2023scanet}}& 89.46 & 91.71 &89.78 &90.43 &90.35 \\
\hline 
\multicolumn{1}{c}{FTNet(Ours)}& 90.06 &\textbf{93.41} &\textbf{93.47} &\textbf{92.75} &\textbf{92.42} \\
\bottomrule 
\end{tabular}
\caption{Comparison with state-of-the-art methods on occluded images from the SFPD dataset. (h\_occ:hand\_occlusion, e\_glasses:eye\_glasses and m\_mask:mouth\_mask)}
\label{tab_2.2}
\end{table}

\textcolor{black}{As Tab.\ref{tab_2} shows, our FTNet achieves better performance on non-occluded image dataset (average F1 1.96 \% better than the second best method \citet{kim2022end}, and 2.13 \% better than the third best method \citet{lin2021roi}). As shown in Tab.\ref{tab_2}, \citet{lin2021roi} outperforms our model in skin parsing. However, as shown in Tab.\ref{tab_2.2}, it falls far short of our model in mouth mask occlusion (by 8.5\%). This is because \citet{lin2021roi} perform the center as the origin of the transform. As in the illumination theory, the radiation range of the surface light source is better than that of the point light source, so that the performance of their model will greatly reduce when the center is occluded, such as face mask occlusion. However, under such occlusion, our method can utilize the homogeneous tanh-transforms of the four corners to alleviate the adverse effect of occlusion on face parsing.}

\textcolor{black}{In terms of the occlusions, Tab.\ref{tab_2.2} also demonstrates that our proposed FTNet (overall 92.42\% ) exhibits superior robustness in occlusion scenarios, outperforming \citet{li2023mask} (overall 90.61\%, the second-placed method) by a margin of 1.81\%. This further demonstrate the effectiveness of our proposed method on occlusions.}

\begin{figure}[!htbp]
	\centering
	\includegraphics[width=5in]{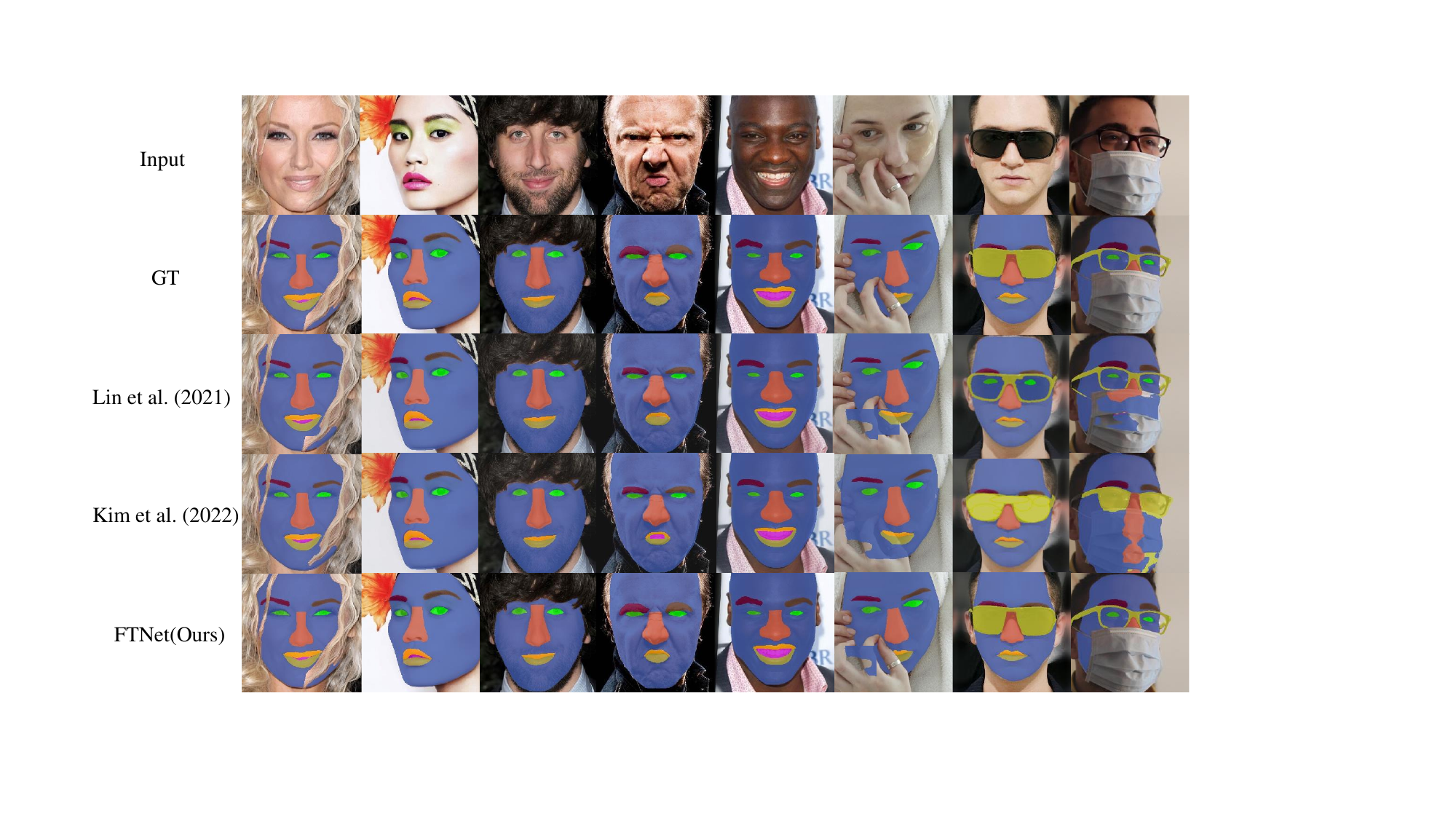}
	\caption{Qualitative comparisons with state-of-the-art methods.}
	\label{fig_x}
\end{figure}

The quantitative analysis of these methods is illustrated in Fig. \ref{fig_x}.
The testing results on SFPD datasets (non-occluded images and occluded images) show that our method outperforms all state-of-the-art methods. Such performance gains are particularly impressive considering that improvement on our dataset is very challenging.

\subsection{Ablation Study}
To demonstrate how each component in our FTNet contributes to the performance, a series of ablation experiments are conducted on SFPD using F1 metric. As the proposed homogeneous Tanh-transforms, i.e., four-point transform, provides a more general and powerful form of image representation, we first quantity the effectiveness of each essential component of FTNet (Section \ref{sec:4.3.1}), and then perform in-depth analyses of homogeneous tanh-transforms (Section \ref{sec:4.3.2}). This would better verify our main points and coincide with the experiments in prior conference papers. The training and evaluation followed the same protocol as in Section \ref{subsec:4.1}.

\subsubsection{Ablation Study for Each Component of FTNet}
\label{sec:4.3.1}
Table \ref{tab_4} shows the evaluation of our FTNet compared to ablated version without certain key components, including homogeneous Tanh-transform, four-point block, occlusion-aware loss. $V_1$ denotes the ablated version without homogeneous tanh-transforms, $V_2$ denotes the ablated version without four-point block, $V_3$ denotes the ablated version without occlusion-aware loss and $V_0$ denotes the full FTNet. All the variants are retrained independently with their specific architectures. we report their average performance on face parts and occlusions (i.e., sunglasses, eyeglasses, face mask) respectively. 

\begin{table}[!htbp] 
\centering
\begin{tabular}{@{}cccccc@{}} 
\toprule 
\multicolumn{3}{c}{Network}& F1 for face parts & F1 for occlusions\\  
\hline 
\multicolumn{3}{c}{\makecell[c]{full FTNet $V_0$ \\ (homogeneous tanh-transforms \\ + four-point block \\ +  occlusion-aware loss)}}& 92.76 &93.21 \\   
\multicolumn{3}{c}{\makecell[c]{ablated FTNet $V_1$ \\(four-point block \\+ occlusion-aware loss)}}&85.46 &83.43  \\
\multicolumn{3}{c}{\makecell[c]{ablated FTNet $V_2$ \\(homogeneous tanh-transforms \\+ occlusion-aware loss)}}&86.71 &88.31\\
\multicolumn{3}{c}{\makecell[c]{ablated FTNet $V_3$ \\(homogeneous tanh-transforms \\ + four-point block)}}&85.28 &85.16 \\  
\bottomrule 
\end{tabular}
\caption{Result on Different Transform. F1 scores are represented in percentage.}
\label{tab_3}
\end{table}

\emph{Homogeneous tanh-transforms.}Consider the first rows in Table \ref{tab_3}, we can find that the homogeneous tanh-transforms provides substantial performance gain. Note that the ablated network uses the original image as the input directly instead of four-point transformed image.This demonstrates the benefit of four-point transform and Tanh-polar representation in occluded face parsing. 

\emph{Four-point block.} Instead of original block in backbone, four-point block can also bring a performance gain (85.28-92.76 for face parts, 85.16-93.21 for occlusion). This suggests that feature fusion of Tanh-Cartesian and Tanh-polar coordinate systems enables comprehensive understanding of facial occlusion semantics.

\emph{Occlusion-aware loss.} Comparing the performance of our full FTNet and the ablated version without occlusion-aware loss, we can conclude that the occlusion-aware loss boosts performance, as the occluded face margin can be con-strained. This also provides a new glimpse into the learning objective and constraint for occlusion handling.

\subsubsection{In-depth Analyses for Homogeneous Tanh-transforms}
\label{sec:4.3.2}

\begin{table}[!htbp] 
\centering
\begin{tabular}{ccccc} 
\toprule 
\multicolumn{3}{c}{Transform}& face parts &sheltered areas\\  
\hline 
\multicolumn{3}{c}{None}& 80.75 &73.79 \\   
\multicolumn{3}{c}{Resize}&74.56 &70.39  \\
\multicolumn{3}{c}{Tanh-Cartesian transform}&82.66 &79.03\\
\multicolumn{3}{c}{Tanh-Polar transform}&85.75 &81.37 \\
\multicolumn{3}{c}{Four point transform (ours)}& \textbf{92.76} &\textbf{93.21} \\   
\bottomrule 
\end{tabular}
\caption{Result on Different Transform. F1 scores are represented in percentage.}
\label{tab_4}
\end{table}
 In above section, we investigate the necessity of comprehensive exploring homogeneous tanh-transforms and the other component of FTNet. Next,we discuss the effectiveness of the homogeneous Tanh-transforms representation. Concretely, we studied three related transforms, as list in Tab. \ref{tab_4}: 1) Resize, 2) Tanh-Cartesian transform by \citep{lin2019face}, 3) Tanh-Polar transform by \citep{lin2021roi}, 4) No transform. All transforms are applied to the dataset before extracting features to constitute the variant model. All five variant models use same backbone and other training parameters. We also evaluate these five models on masked images, then check the dice loss, our loss and the F1 score for face parts or occlusions. We can draw a main conclusion that our homogeneous tanh-transforms, i.e., four-point trans-form, helps with semantic learning of occluded faces as our full model is significantly better than the one without four-point transform which further proves the generality of our method.

\subsection{Qualitative Results}

\begin{figure}[!htbp]
	\centering
	\includegraphics[width=5in]{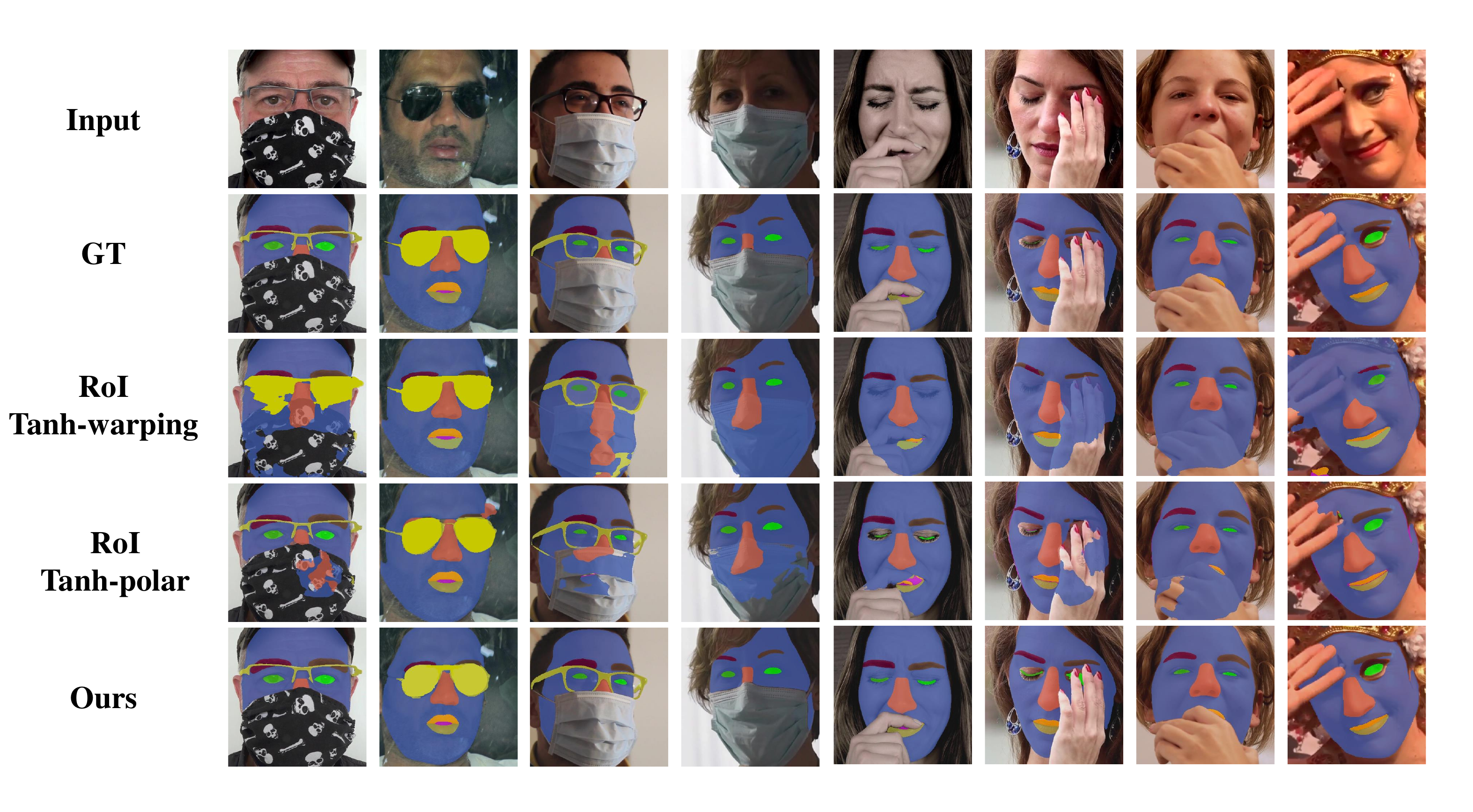}
	\caption{Qualitative comparisons in cropped images between previous methods RoI Tanh-warping \citep{lin2019face}, RoI Tanh-polar \citep{lin2021roi} and ours on the SFPD. The proposed four-point transform addressed the typical occlusion, including sunglasses, eyeglasses, hand and face mask.}
	\label{fig_5}
\end{figure}
\begin{figure}[!htbp]
	\centering
	\includegraphics[width=5in]{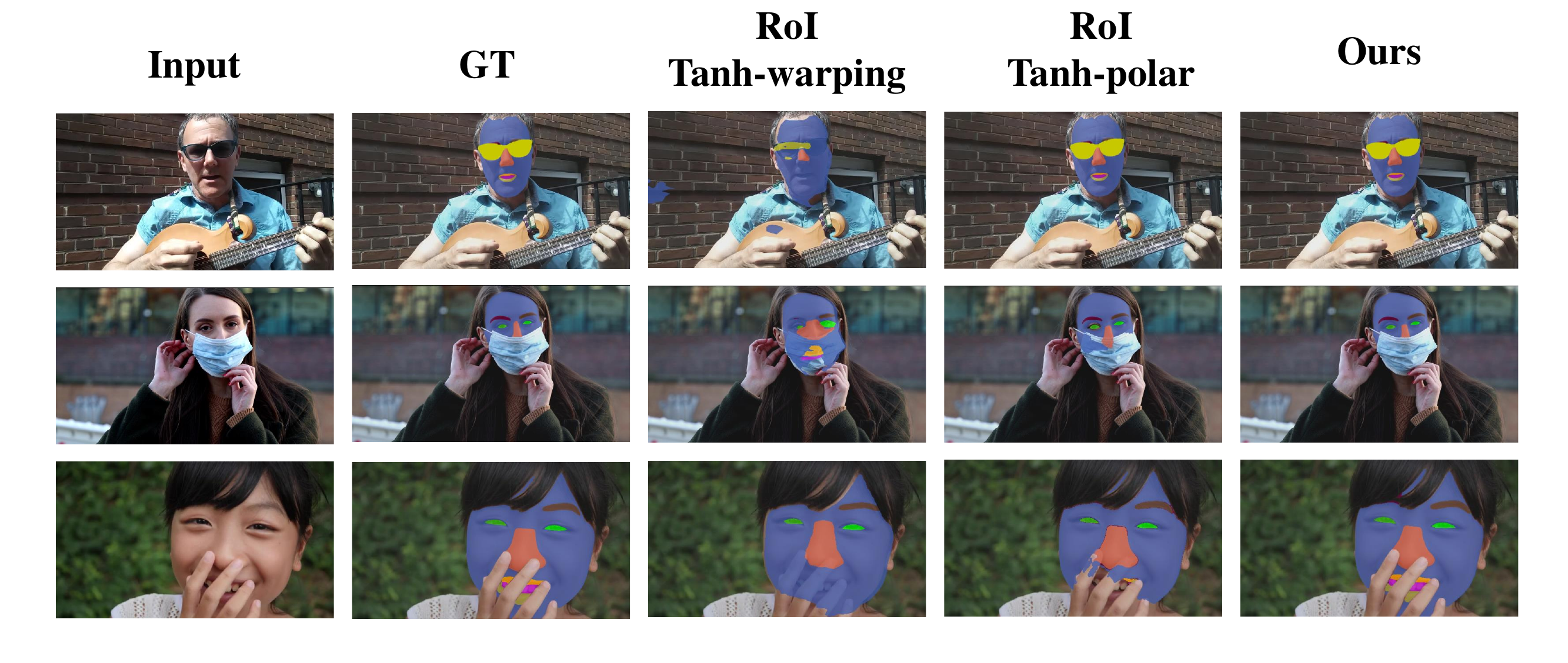}
	\caption{Qualitative comparisons in wild scenes between previous methods RoI Tanh-warping \citep{lin2019face}, RoI Tanh-polar \citep{lin2021roi} and ours.}
	\label{fig_6}
\end{figure}

\begin{figure}[!htbp]
	\centering
	\includegraphics[width=5in]{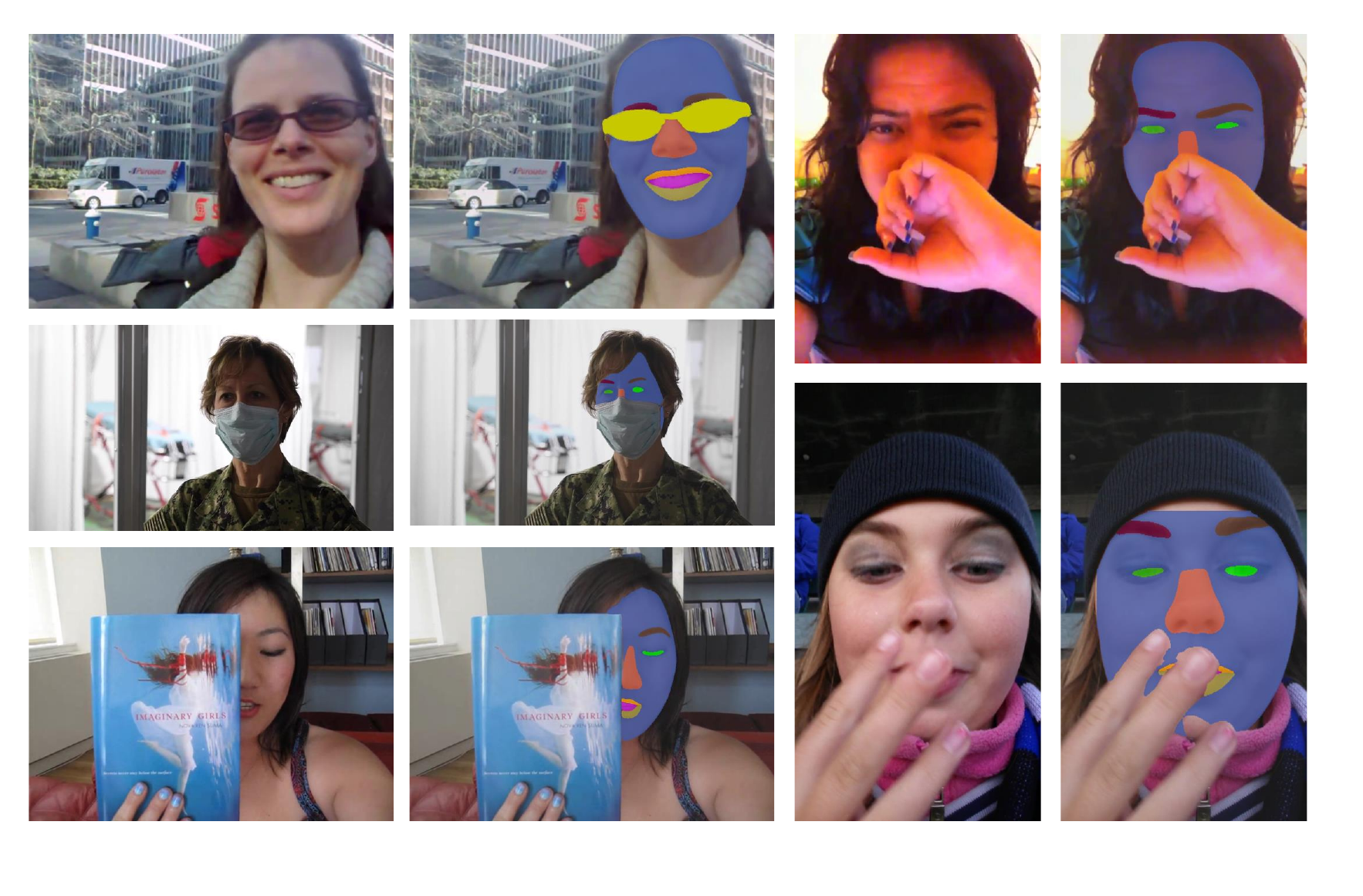}
	\caption{Qualitative results on other challenging images with complicate occlusions. }
	\label{fig_7}
\end{figure}
Finally, we compare the proposed FTNet with two most related competitors (i.e., RoI Tanh-warping \citep{lin2019face} and RoI Tanh-polar \citep{lin2021roi}) on SFPD. Some qualitative comparison results on the cropped images are depicted in Fig.\ref{fig_5}. We can see that our approaches output more precise parsing results than other competitors under the occlusions, including sunglasses, eyeglasses, hand and face mask. Neither RoI Tanh-warping \citep{lin2019face} nor RoI Tanh-polar \citep{lin2021roi} can precisely parse face parts under all occlusion conditions. The performance of these two methods can be accepted under sunglasses and eyeglasses (2nd column), because this kind of occlusion differ greatly from the skin color and can be easily distinguished in most cases. However, when the occlusion prone to skin color confusion occurs, such as face masks and hands, these methods are prone to parsing errors and far inferior to our proposed method. In addition, with its better leveraging of illumination theory, our four-point transform eliminates the interference from the occlusion and gets more robust results.

The qualitative comparisons on the complete scene image are shown in Fig. \ref{fig_6} and Fig. \ref{fig_7} gives some other challenging case, where our FTNet still correctly recognizes the confusing parts of the person under occlusion. The qualitative results show that the significant improvement brought by our FTNet in parsing occluded faces. 

\subsection{Failure Case Analysis}
\begin{figure}[!htbp]
	\centering
	\includegraphics[width=5in]{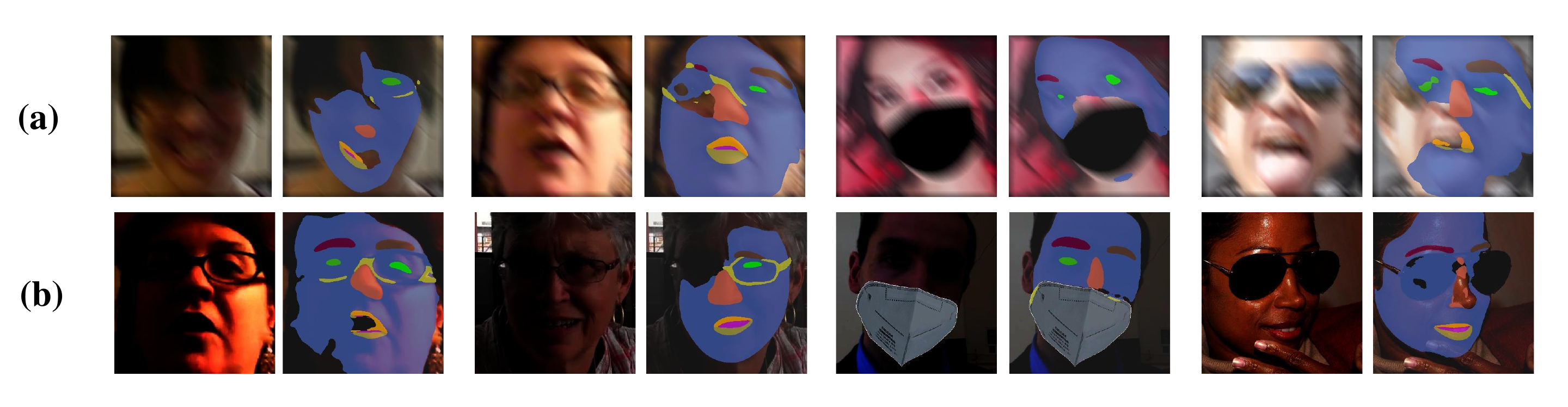}
	\caption{Visualizations of typical failure cases on SFPD test set, including (a) low-quality images, (b) low-light images.}
	\label{fig_8}
\end{figure}
To give a deeper insight into our method, we present two representative failure cases in Fig. \ref{fig_8}. As seen our proposed model face difficulties with low quality images or low-light occluded scenes. In the future, we will therefore focus on addressing these issues.

\section{Conclusion}
\label{sec:conclusion}
In this paper, we propose a novel occlusion-aware network combined with homogeneous tanh-transforms for occluded face parsing, namely, four-point trans-form neural network FTNet. We use four-point transform to warp the image to four Tanh-polar coordinate systems, which provides a high anti-occultation ability. Our occlusion-aware network utilizes receptive fields of different shapes for warped images, while compressing more information of surrounding context. Moreover, an occlusion-aware loss also enhances the robustness of the model to occluded face parsing. Ablation studies show the effectiveness of four-point transform and our proposed method. The superior performance show that we have solved the dilemma of face parsing on occluded faces in most wild scenes and our method is highly capable on face parsing tasks with face coverings. In the future work, we would like to focus on the addressing difficult case, including low-quality and low-light images, as well as apply our method into other occluded image segmentation tasks, such as human parsing and object segmentation.




\bibliographystyle{elsarticle-harv} 
\bibliography{cas-refs}





\end{document}